\pdfoutput=1

\documentclass[11pt]{article}

\usepackage[preprint]{acl}

\usepackage{times}
\usepackage{latexsym}

\usepackage[T1]{fontenc}

\usepackage[utf8]{inputenc}

\usepackage{microtype}

\usepackage{inconsolata}

\usepackage{graphicx}
\usepackage{xcolor} 

\usepackage{amsfonts}

\newcommand{\Singular}{\textit{S}}
\usepackage{booktabs}
\usepackage{multirow}
\newcommand{\our}{B-LoRA-XS}
\newcommand{\tB}{B}
\usepackage{mathabx}

\usepackage{float}

\title{Minimal Ranks, Maximum Confidence: \\ Parameter-efficient Uncertainty Quantification for LoRA}

\author{Patryk Marsza{\l}ek\thanks{Denotes primary contributors.} \quad Klaudia Bałazy \quad Jacek Tabor \quad Tomasz Ku\'smierczyk\footnotemark[1]\thanks{Correspondence: \href{mailto:t.kusmierczyk@uj.edu.pl}{t.kusmierczyk@uj.edu.pl}}
\\ \\ Jagiellonian University
}

\begin{document}
\maketitle
\begin{abstract}
Low-Rank Adaptation (LoRA) enables parameter-efficient fine-tuning of large language models by decomposing weight updates into low-rank matrices, significantly reducing storage and computational overhead. While effective, standard LoRA lacks mechanisms for uncertainty quantification, leading to overconfident and poorly calibrated models. Bayesian variants of LoRA address this limitation, but at the cost of a significantly increased number of trainable parameters, partially offsetting the original efficiency gains. Additionally, these models are harder to train and may suffer from unstable convergence.  
In this work, we propose a novel parameter-efficient Bayesian LoRA via subspace inference, demonstrating that effective uncertainty quantification can be achieved in very low-dimensional parameter spaces. The proposed method achieves strong performance with improved calibration and generalization while maintaining computational efficiency. Our empirical findings show that, with the appropriate projection of the weight space: (1) uncertainty can be effectively modeled in a low-dimensional space, and (2) weight covariances exhibit low ranks.
\end{abstract}

\section{Introduction}

LoRA (Low-Rank Adaptation)~\cite{hu2021lora} reduces computational overhead by decomposing the update weights of pre-trained models into low-rank matrices, enabling efficient adaptation to downstream tasks.  
Minimizing the number of trainable parameters reduces memory and storage requirements, making large-scale model adaptation feasible. Reducing computational overhead speeds up training time and makes adaptation possible in resource-constrained settings.  

Unlike pre-trained models, which are relatively well-calibrated \citep{openai2023gpt4}, fine-tuned large models (e.g., LLMs) often become overconfident and poorly calibrated~\citep{jiang2021can,tian2023just,xiao2022uncertainty,he2023preserving}, especially when trained on limited data. This hinders their usability for applications where uncertainty-aware decisions are performed. 

\begin{figure}[t]
    \centering
    \includegraphics[width=1.0\linewidth]{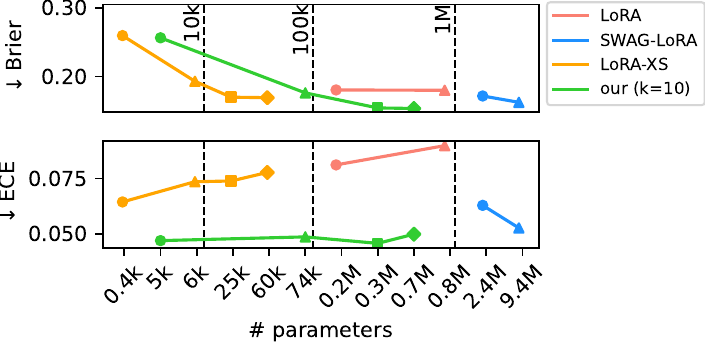}
    \caption{
Performance averaged over multiple GLUE datasets (individual results in Fig.~\ref{fig:lorar}). Our method achieves superior calibration (ECE) and competitive predictive performance (Brier) while maintaining computational efficiency. For example, at $r=8$ ($\blacktriangleup$), we reduce ECE by half with only 1/10th LoRA parameters.  
    }
    \label{fig:teaser}
\end{figure}

\begin{figure*}[t]
    \centering
    \begin{minipage}{0.7\textwidth}
    \includegraphics[width=1.0\linewidth]{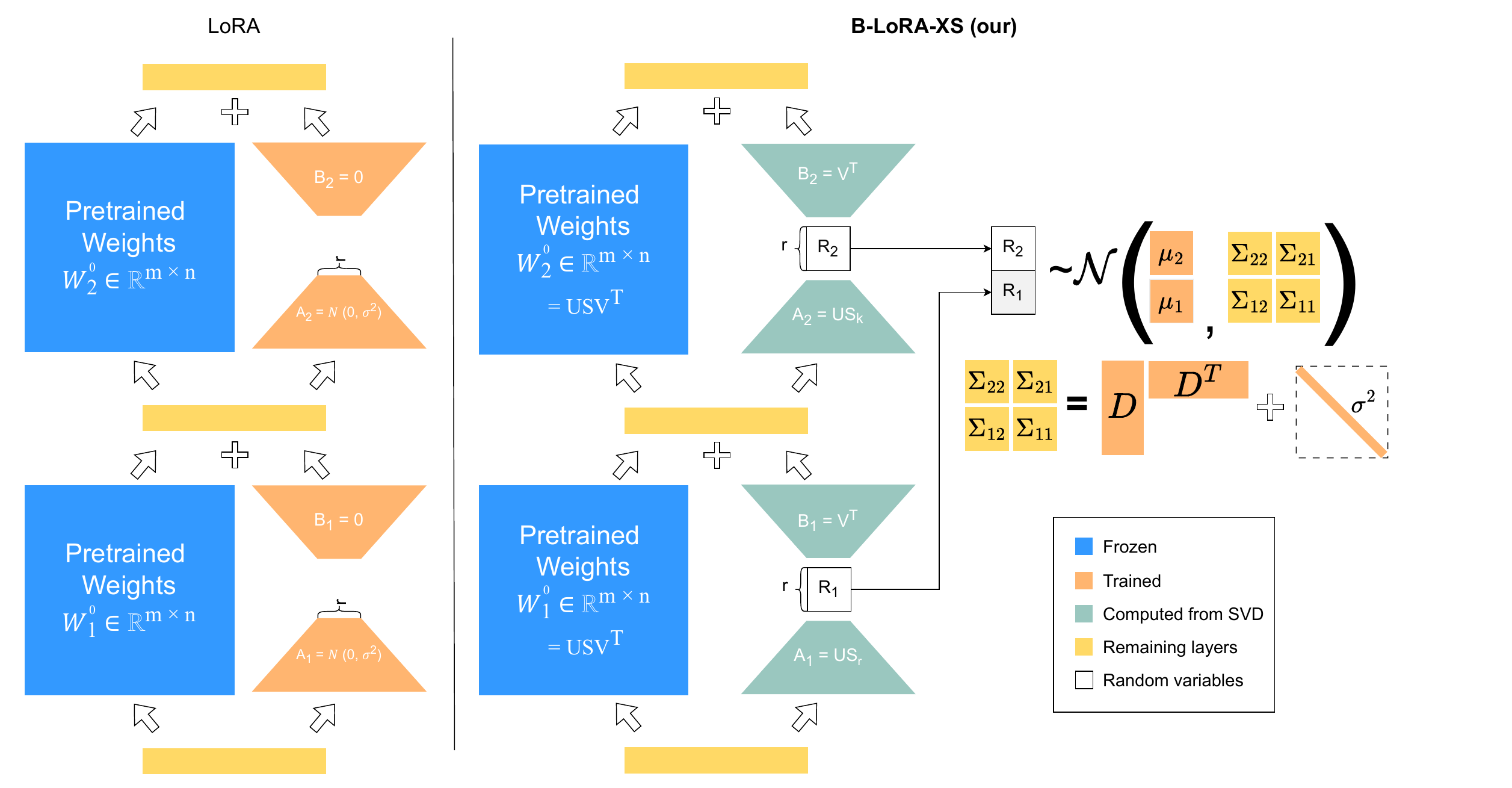}
    \end{minipage}%
    \hfill
    \begin{minipage}{0.29\textwidth}
    \resizebox{1.0\textwidth}{!}{
    \begin{tabular}{llrrr}
        \toprule
         & \textbf{Method} & \textbf{$r$} & $k$ & \textbf{\# Parameters} \\
        \cmidrule{1-5}
        \multirow{4}{*}{\rotatebox{90}{Standard}} & LoRA & 2 & - & 0.2M \\
        & LoRA & 8 & - & 0.8M \\
        \cmidrule{2-5}
        & LoRA-XS & 8 & - & 6k \\
        & LoRA-XS & 25 & - & 60k \\
        \midrule
        \multirow{6}{*}{\rotatebox{90}{Bayesian}} & SWAG-LoRA & 2 & 10 & 2.4M \\
        & SWAG-LoRA & 8 & 10 & 9.4M \\
        & SWAG-LoRA & 8 & 5 & 5.5M \\
        \cmidrule{2-5}
        & \emph{\our} & 8 & 10 & 74k \\
        & \emph{\our} & 25 & 10 & 0.7M \\
        & \emph{\our} & 25 & 5 & 0.4M \\
        \bottomrule
    \end{tabular}

        }
    \end{minipage}
    \caption{
    \textbf{(left):}~Weight-adaptation approaches: LoRA vs.\ \our.
    As indicated by the color coding, some parameters remain frozen (\emph{blue}), others are trained (\emph{orange}) or obtained via SVD (\emph{green}).
    \textbf{(right):}~Number of trainable parameters per method. XS variants remain computationally competitive even for ranks as large as $r=25$.   
    }
    \label{fig:method}
\end{figure*}

Bayesian treatment is then frequently proposed to address overconfidence in neural networks~\citep{blundell2015weight,kristiadi2020being,aitchison2021deep,izmailov2021bayesian}. Consequently, recently proposed Bayesian variants of LoRA~\citep{onal2024gaussian,robeyns2024laplaceLora,doan2025bayesianlowranklearningbella} address the aforementioned challenges by introducing uncertainty estimation directly into the fine-tuning process. During training, these models continuously adjust both the mean and covariance of fine-tuned parameters to achieve better generalization and uncertainty quantification.  

Learning the posterior covariance matrix is necessary for modeling epistemic uncertainty. However, its size grows quadratically with the number of parameters, which can easily cancel out the benefits of LoRA, in addition to making learning significantly harder. Using low-rank, Kronecker-factored, or diagonal-only covariances partially alleviates the problem, but as we demonstrate in Sec.~\ref{sec:experiments}, this comes at the cost of results quality loss. Furthermore, even at rank = 2, the number of trainable parameters is quadrupled compared to vanilla LoRA. This creates a need for an alternative approach that retains covariance modeling capacity while reducing the number of required parameters.  

We propose
a method that learns Bayesian posteriors for weights projected onto a low-dimensional manifold, hence maintaining parameter efficiency. The thoughtfully selected projection allows for the effective representation of the covariances between weights through covariances between representations in the lower-dimensional space. In this design, we follow the work of \citet{balazy2024lora}, who recently proposed a strategy for finding such projections with SVD. We prove that they are \emph{effective for learning Bayesian models as well}.

Operating in such a reduced parameter space significantly improves the feasibility of Bayesian inference. 
We show that 
correlations between weights can be represented very efficiently -- unlike in the original weight space, we can use covariance matrices with ranks as low as $k=2$. Thanks to the low number of parameters, training is also more stable. Finally, the method achieves superior calibration and accuracy at low budgets (e.g., see Fig.~\ref{fig:teaser}). 

A key contribution of our work lies in introducing Bayesian learning within a low-rank projected subspace derived from pre-trained weights. While the individual components of our method, such as SVD projections and Bayesian inference, are established techniques, their synergistic application to learn Bayesian posteriors within a compressed subspace constitutes a meaningful conceptual innovation. Our approach enables uncertainty-aware fine-tuning with strong parameter efficiency, yielding improved uncertainty quantification with minimal computational overhead.

In the Appendix, we supplement the results presented in the paper with a discussion of related work, a detailed overview of the experimental setup, and the exact numeric values for the figures in the main text. 

The source code is available online\footnote{Source code: \url{https://github.com/gmum/b-lora-xs}}.

\begin{figure*}[ht]
    \centering
    \begin{minipage}{0.925\textwidth}
    {\scriptsize
    \setlength{\baselineskip}{0.0em} 
    \makebox[0.95\textwidth]{CoLA} 
    \includegraphics[width=0.9975\textwidth]{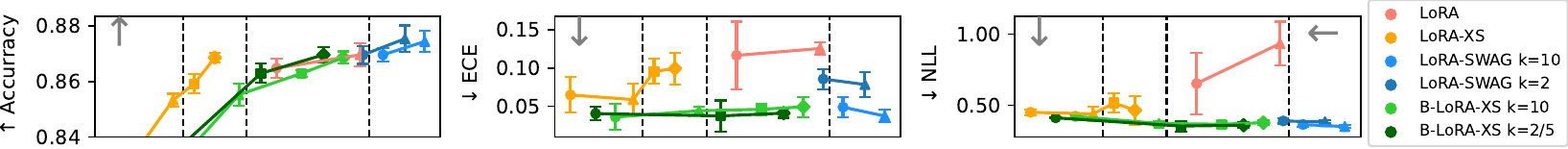} 
    
    \makebox[0.95\textwidth]{MRPC}  
    \includegraphics[width=0.935\textwidth]{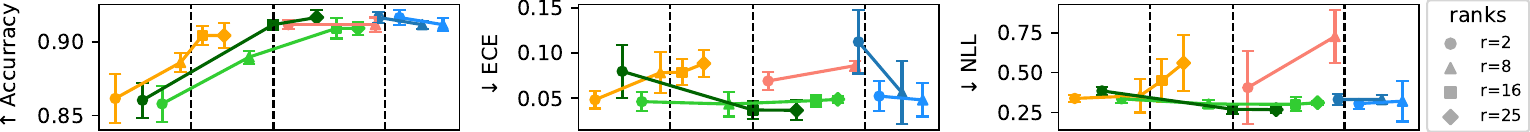} 
    
    \makebox[0.95\textwidth]{RTE}  
    \includegraphics[width=0.868\textwidth]{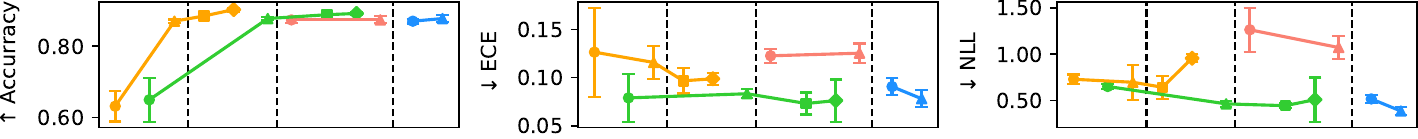} 

    \makebox[0.95\textwidth]{SST-2} 
    \includegraphics[width=0.878\textwidth]{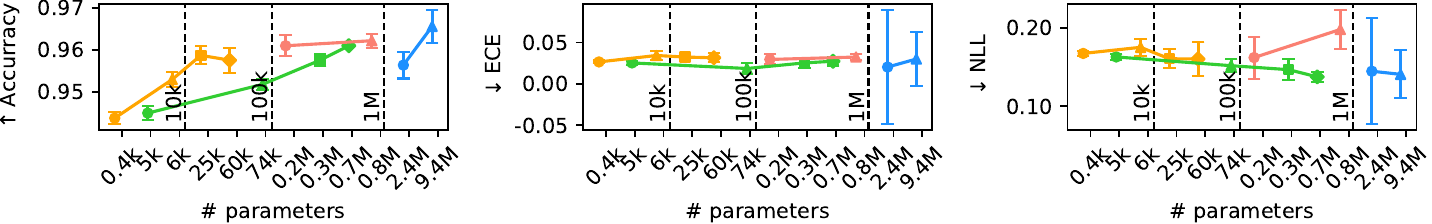}  

    }
    \end{minipage}%
    \\
    \caption{
    Median$\pm$s.d. accuracy (left), ECE (middle), and NLL (right) on 4 GLUE tasks (rows) vs. total parameter count for several methods and varying ranks $r$. \our{} (our) achieves the accuracy and the calibration of SWAG-LoRA while using significantly fewer parameters than LoRA.
    See Fig.~\ref{fig:teaser} for averaged results. {The arrows in the figure indicate the direction of improvement ("towards better") in the standard manner.
    The exact numeric values underlying the plots are reported in Tables~\ref{tab:num_cola}-\ref{tab:num_sst2} in the Appendix.
    }
    }
    \label{fig:lorar}
\end{figure*}

\section{Method: \our}
\label{sec:method}

\textbf{LoRA} fine-tunes large pre-trained models by learning low-rank weight updates $\Delta W$ instead of training the weights $W$ directly. For a pre-trained parameter matrix \(W^0 \in \mathbb{R}^{m \times n}\) that is kept fixed, LoRA learns a rank-\(r\) update \(\Delta W = AB\), where \(A \in \mathbb{R}^{m \times r}\) and \(B \in \mathbb{R}^{r \times n}\) have far fewer parameters. The effective weight is then:
$
W = W^0 + \Delta W = W^0 + AB,
$
where only \(A\) and \(B\) are trained. 
LoRA is then typically applied jointly for multiple layers $l$, yielding a set of updates $\{\Delta W_{l}\}$.
Then, Bayesian treatment of LoRA can improve its calibration and uncertainty quantification.

\textbf{Bayesian treatment} of a neural network involves finding the posterior \(p(\theta \mid \mathcal{D})\) given training data \(\mathcal{D}\). 
By Bayes' theorem:
$
p(\theta \mid \mathcal{D}) \;=\; \frac{p(\mathcal{D} \mid \theta)\,p(\theta)}{p(\mathcal{D})},
$
where $\theta$ represents the model's parameters (i.e., weights) considered random variables. 
Specifically, for the Bayesian LoRA setting, $\theta$ denotes a set of the \emph{learned model updates}, while the remaining \emph{frozen} weights are hidden inside the model likelihood, given by $p(\mathcal{D} \mid \theta) = \prod_{i \in [\mathcal{D}]} p(y_i | x_i, \theta)$.
The learned posterior allows Bayesian model averaging at inference as:
$
p(y_* \mid x_*, \mathcal{D}) \;=\; \int p(y_* \mid x_*, \theta)\,p(\theta \mid \mathcal{D})\,d\theta \approx 
\frac{1}{S} \sum_{\theta \sim p(\theta|D)} p(y_* \mid x_*, \theta).
$

\textbf{Bayesian LoRAs} obtain the posterior for $\{\Delta W_{l}\}$ through the learned posterior for $\theta = \cup_l \{ A_{l} \cup B_{l} \}$, where $l$ indexes the weight updates (layers). The posterior itself is approximated either using a set of particles or a closed-form distribution. Due to its superior performance, we rely on the latter and assume $p(\theta | \mathcal{D}) \approx \mathcal{N}(\mu, \Sigma)$, where $\mu$ is the vector of means (of size equal to the number of learned parameters) and $\Sigma$ is the covariance matrix, whose size grows quadratically with the total number of parameters. Notably, we aim to \emph{model cross-layer interdependencies}, requiring covariance estimation also across weights in different layers $\{l\}$. This however results in an impractically large number of parameters.
Consequently, we explore methods to reduce this cost by representing distributions $p(\{\Delta W_l\} | \mathcal{D})$ differently, e.g., using SVD-based projections.

\textbf{In LoRA-XS}~\cite{balazy2024lora}, the adaptation matrices $A$ and $B$ are initialized using the truncated SVD of the corresponding pre-trained weight matrices $W^0$. 
This initialization captures the most informative singular components of the original weights. Under the assumption that the fine-tuned task is similar to the original task, these projections retain the functional properties also for downstream adaptations. 
LoRA-XS then \emph{freezes} $A$ and $B$ and inserts a small \emph{trainable} matrix $R \in \mathcal{R}^{r \times r}$ between them, reducing the number of trainable parameters to $r^2$ ($r^2 \ll (n + m) \cdot r$) per weight matrix.
Then, the fine-tuning update is:
$
h = x W^0 + x \Delta W = x W^0 + x A R B,
$
where $A \in \mathbb{R}^{m \times r}$ and $B \in \mathbb{R}^{r \times n}$ are low-rank matrices obtained from the truncated SVD of $W^0$, specifically $A = U_r \Singular_r$ and $B = V_r^T$.

{\our}, proposed in this paper, leverages the frozen projections $A$ and $B$ for effective and efficient Bayesian learning. 
{Its core idea is to apply the Bayesian treatment in the extremely compressed parameter space given by SVD-based projections, making Bayesian inference tractable and highly efficient. Although this specific idea is novel, 
it can be related to the framework of subspace Bayesian Inference.

\textbf{Subspace Inference} (SI)~\citep{izmailov2020subspace} was proposed as a remedy for the intractability of full-dimensional posteriors in modern networks. \emph{Low-dimensional affine manifold}, carefully centred on a well-trained solution, already contains a rich family of high-performing weight vectors. Performing Bayesian integration restricted to that manifold restores calibrated uncertainty without revisiting the entire parameter space.
In particular,
given a well-trained reference point $\bar w\in\mathbb{R}^{d}$ and $K\ll d$ orthonormal basis vectors-e.g. the leading PCA directions of an SGD trajectory - the \emph{learning subspace} is
$
\mathcal{S}:=\bigl\{ w=\bar w+Pz  \bigm|  z\in\mathbb{R}^{K}\bigr\}$, $P\in\mathbb{R}^{d\times K},
$
where Bayesian parameters are the low-dimensional representations $\theta = \{z\}$.

\textbf{\our} defines a \emph{layer-local} manifold that leaves the pretrained backbone untouched. For each frozen weight matrix $W^{0}_{l}$, an SVD
$W^{0}_{l}=U_{l}S_{l}V_{l}^T$
is computed once, and the top $r$ singular directions yield fixed projectors $A_{l}=U_{l,r} S_{l,r}$ and $B_{l}=V_{l,r}^T$. Note that, whereas SI \emph{learns} $P$ from the training dynamics, B-LoRA-XS \emph{projects} onto directions already favored by the pretrained backbone. 
Then, 
all task-specific variability is captured by these square adapters $R_{l}\in\mathbb{R}^{r\times r}$.

Vectorising and stacking these layer-projections defines
$
w = \bar{w}+P_{\tB} z_{\tB}
$,
$
P_{\tB}=\mathrm{blockdiag}\bigl(B_{l}^T\otimes A_{l}\bigr),
$
where $w = [ \mathrm{vec}(W_1)^T,  \ldots,  \mathrm{vec}(W_l)^T,  \ldots]^T$, $\bar{w} = [ \mathrm{vec}(W_1^0)^T,  \ldots,  \mathrm{vec}(W_l^0)^T,  \ldots]^T$, and $P_{\tB}$ is a block-diagonal matrix with blocks $B_{l}^T \otimes A_{l}$, one for each layer~$l$.

Thus B-LoRA-XS explores an affine subspace 
$\mathcal{S}_{\tB}:=\{w = w^{0}+P_{\tB}z_{\tB}| z_{\tB}\in\mathbb{R}^{\sum_{l} r^{2}}\},$
 whose dimension scales with $r^{2}$ per adapted layer.

In Sec.~\ref{sec:experiments}, we empirically demonstrate that $A_l$ and $B_l$, obtained from the SVD of the pre-trained weights, are not only effective for point-wise fine-tuning but also enable effective uncertainty quantification for $\{\Delta W_l\}$ through modeling covariances for $\{ R_l \}$.
Although we never compute it explicitly, the covariance matrix for individual $\Delta W$ is expressed as $\Sigma_{\Delta W_{l}} = (B_{l}^T  \otimes A_{l})\Sigma_{R_{l}}(B_{l}^T \otimes A_{l})^T,$, where $\Sigma_R$ is the (intra-layer) covariance matrix for $R$ and $\otimes$ denotes the Kronecker product.

In practice, we simply learn the \emph{joint posterior} $p(\theta = \cup_l R_l | \mathcal{D}) \approx \mathcal{N}(\mu, \Sigma)$ (only) for the inner matrices $R$. The covariance matrix $\Sigma$ captures both inter-layer and intra-layer dependencies, allowing the model to learn complex relationships.
At inference, similar to LoRA, we use samples of $R$ along with the respective projections $A$ and $B$ to obtain $h$, as realized through samples of $\Delta W$, however without ever computing it explicitly.

The parameters $\mu$ and $\Sigma$ are learned efficiently using \textbf{SWAG}~\cite{NEURIPS2019_118921ef} (though Variational Inference or the Laplace approximation could also be used). After a burn-in phase (a fixed 10 or 25 epochs) of the gradient-based optimization, the algorithm maintains $\hat \mu$ -- a running average of $\theta$ -- along with $k$ vectors of differences $\hat D_{{last}} = \theta_{{last}} - \hat \mu$ for the last $k$ values of $\theta$, and a running average of $\theta^2$. Based on these averages, we estimate the variances $\hat \sigma^2$ for individual parameters and approximate the covariance as
$
\hat \Sigma \approx \frac{1}{2}(\hat D \cdot \hat D^T + {diag}(\hat \sigma^2)),
$
which constitutes a rank-$k$ approximation to the covariance matrix.

We illustrate \our{}   method 
in Fig.~\ref{fig:method}.
Our method uses the total of $|\theta| \cdot (k+2)$ parameters, where $|\theta| = \sum_l r^2$.

\begin{figure*}[!t]
    \begin{minipage}{1.0\textwidth}
    \centering
    {\scriptsize 
    CoLA\\
    \hspace{0.625cm}
    \includegraphics[width=0.93\linewidth]{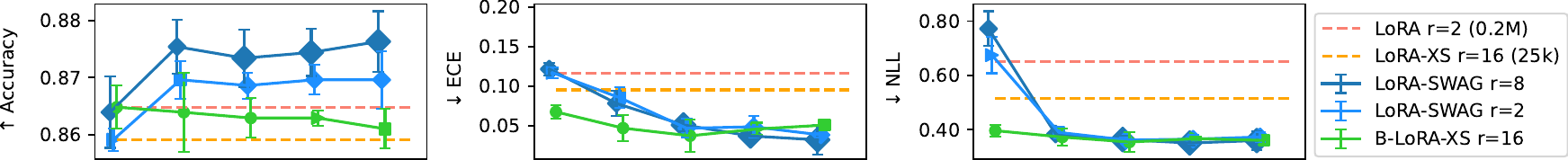} \\
    \vspace{0.25cm}
    MRPC\\
    \includegraphics[width=0.88\linewidth]{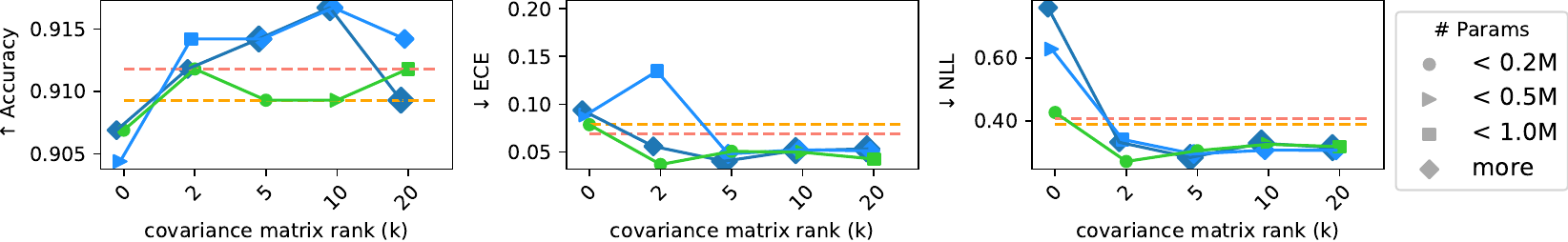}\\
    }    
    \end{minipage}
    \caption{
Impact of the posterior covariance matrix rank ($k=0$ indicates the case with no off-diagonal elements) for CoLA (top) and MRPC (bottom). For brevity, confidence bars ($\pm$ standard deviation) are omitted for MRPC. The colored lines represent non-Bayesian baselines (e.g., standard LoRA or LoRA-XS at a given rank $r$).   
    {    The exact numeric values underlying the plots are reported in Tables~\ref{tab:cov_cola}~and~\ref{tab:cov_mrpc} in the Appendix.}    
    }
    \label{fig:covrank}
\end{figure*}

\begin{figure*}[ht]
    \centering
    {\scriptsize 
    CoLA\\
    \includegraphics[width=0.885\linewidth]{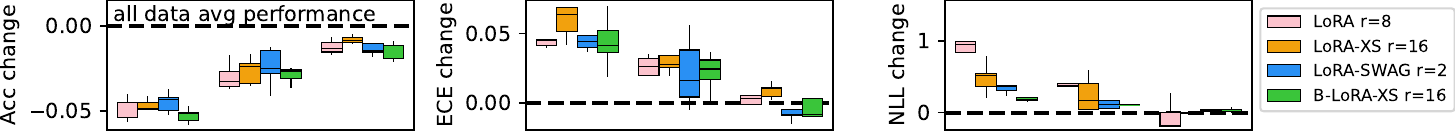} \\
    \vspace{0.25cm}    
    MRPC\\
    \hspace{0.075cm}
    \includegraphics[width=0.88\linewidth]{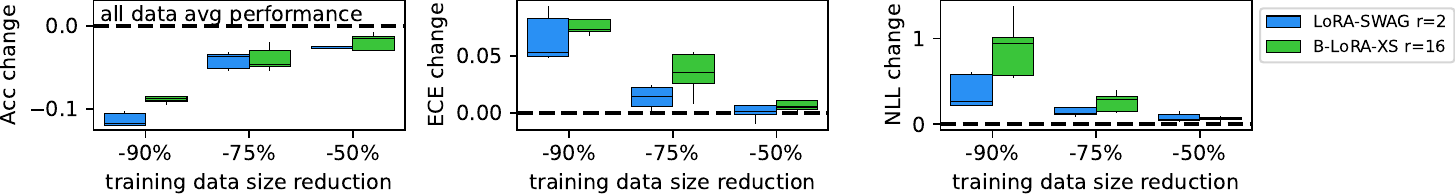}\\
    }
    \caption{Accuracy, ECE and NLL change as the training set is progressively reduced (e.g.\ -90\% means using only 10\% of the data for training). The dashed line marks the model's performance when trained on the full dataset.}
    \label{fig:subsampling}
\end{figure*}

\section{Experiments}
\label{sec:experiments}

\textbf{Setup:}
We performed experimental evaluation on four GLUE tasks~\cite{wang2019} (RTE, MRPC, CoLA, and SST-2) using RoBERTa-large~\cite{liu2019}. We compare our method (\our) against the standard LoRA, LoRA-XS -- a parameter efficient variant, and against SWAG-LoRA~\cite{onal2024gaussian} -- a Bayesian variant.
For LoRA-XS and \our{} we considered ranks $r \in \{2, 8, 16, 25\}$ and for LoRA and SWAG-LoRA due to limited budget we were able to test $r \in \{2, 8\}$. The number of parameters (a \emph{proxy for storage and computation costs}) as a function of ranks $r$ and $k$ is summarized in Fig.~\ref{sec:method}.  
We report accuracy (higher is better), ECE and NLL (lower is better) of median$\pm$s.d. across 5 runs.

\textbf{Performance analysis:} 
Fig.~\ref{fig:lorar} compares accuracy, Expected Calibration Error (ECE), and Negative Log-Likelihood (NLL) against total parameter count across 4 datasets. 
{Our main claim is that \our{} improves overall model performance, with a particular focus on calibration metrics. Indeed, Figure~\ref{fig:lorar} (middle and right) demonstrates that \our{} consistently yields lower ECE and NLL compared to standard LoRA across all parameter scales. Regarding accuracy (Figure~\ref{fig:lorar}: left), while standard LoRA shows marginally better results for a few configurations at moderate parameter scales, the majority of configurations show \our{} matching or exceeding the accuracy of standard LoRA. More importantly, in no setting does standard LoRA significantly outperform \our{} in terms of calibration, which is a primary focus of our work. Bayesian variants, including \our{} and SWAG-LoRA, generally outperform their non-Bayesian counterparts in ECE and NLL. However, our model achieves these strong calibration results with 5–15 times fewer parameters than SWAG-LoRA.}
Moreover,
while SWAG-LoRA sometimes performs well, its results vary significantly between runs. In contrast, \our{} exhibits stable and consistent convergence. Finally, as results for MRPC and CoLA suggest, its performance remains robust across different values of $k$, whereas SWAG-LoRA's ECE deteriorates significantly at $k=2$.

\textbf{Covariance matrix rank analysis:} 
Figure~\ref{fig:covrank} compares the sensitivity of the Bayesian LoRA variants to changes in covariance matrix rank $k$.  
Markers indicate model sizes (e.g., SWAG-LoRA $\gg$ \our). As expected, SWAG-LoRA deteriorates proportionally as rank decreases. On the other hand, \our{} maintains its performance across a wide range of $k$. Significant degradation occurs only when off-diagonal covariance values are entirely ignored (i.e., at $k=0$). Notably, \our{} achieves the best calibration at low ranks, particularly at $k=2$ or $k=5$. This demonstrates that the SVD-based projection effectively disentangles parameters, enabling low-dimensional uncertainty modeling.

\textbf{Data size reduction analysis:}
Figure~\ref{fig:subsampling} compares how accuracy, ECE, and NLL degrade when training data is subsampled.  
All methods predictably lose accuracy as data size decreases, with little difference between the various LoRA-based approaches.  
We conclude that Bayesian learning does not improve robustness in this case.  
However, we note variations across datasets in terms of accuracy. For example, in MRPC, the decline is more pronounced, likely due to the dataset smaller size.

\section{Conclusion}
\our{} addresses the lack of uncertainty quantification in LoRA fine-tuning while maintaining parameter efficiency. It utilizes truncated SVD to project model updates into a lower-dimensional space and leverages the Bayesian framework to enhance uncertainty estimation.

Our method's primary strength lies in its calibration capabilities; it consistently achieves \emph{lower expected calibration error and negative log-likelihood} compared to standard LoRA and LoRA-XS across various parameter scales. While standard LoRA may exhibit marginally better accuracy in a few specific configurations, \our{} generally matches or exceeds its accuracy in most settings, and critically, always provides superior or equal calibration. Compared to the Bayesian LoRA baseline, \our{} matches or surpasses its accuracy and calibration performance while using significantly fewer parameters, exhibiting greater training stability, and relying on simpler, lower-rank covariance representations.


\section*{Limitations}

{\small
{While \our{} demonstrates promising results in parameter-efficient uncertainty quantification, several limitations should be acknowledged. 
First, the effectiveness of \our{} inherently depends on the quality of the initial SVD projection derived from pre-trained weights (as in LoRA-XS). If the principal components of the pre-trained model are not well-aligned with the requirements of a significantly different downstream task, the performance might be suboptimal. 
Second, our method employs SWAG with a low-rank approximation for the covariance matrix. While efficient, this is one specific approach to approximate Bayesian inference. Other techniques (e.g., more sophisticated variational inference methods or different posterior approximations) might yield different trade-offs between performance, calibration, and computational cost, and were not explored in this work.
Third, although \our{} significantly reduces the number of trainable parameters for Bayesian adaptation, the inference process still requires multiple forward passes for sampling, which increases computational cost compared to non-Bayesian LoRA or LoRA-XS. 
Fourth, our empirical validation is conducted on GLUE classification tasks using RoBERTa-Large. The generalizability of \our{}'s benefits to other model architectures, much larger model scales, or different task types (such as text generation or more complex reasoning tasks) warrants further investigation.
Finally, the optimal choice of LoRA rank $r$ and SWAG covariance rank $k$ might require careful tuning for different datasets and models, potentially adding to the practical overhead of applying the method effectively.}
}

\section*{Acknowledgments}

{\small
This research is part of the project No. \textbf{2022/45/P/ST6/02969} co-funded by the National
Science Centre and the European Union Framework Programme for Research and
Innovation Horizon 2020 under the Marie Skłodowska-Curie grant agreement No.
945339. For the purpose of Open Access, the author has applied a CC-BY public copyright
licence to any Author Accepted Manuscript (AAM) version arising from this submission. 
\\
\includegraphics[width=1cm]{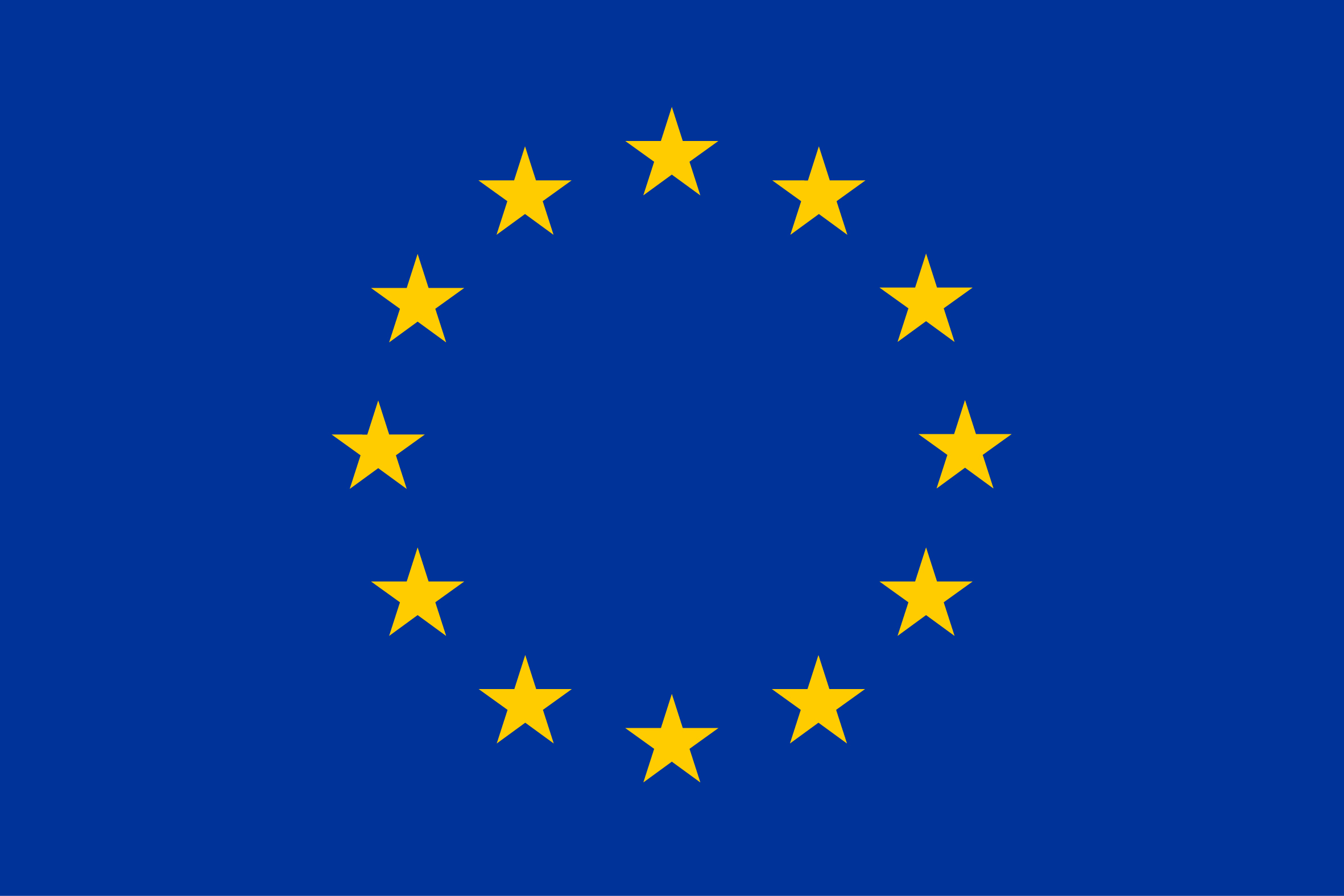} \includegraphics[width=1.9cm]{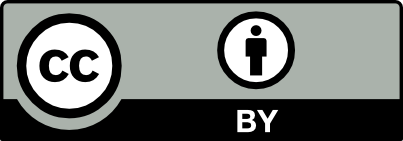}
\\ \\
We gratefully acknowledge Polish high-performance computing infrastructure PLGrid (HPC Center: ACK Cyfronet AGH) for providing computer facilities and support within computational grant no. PLG/2024/017893
\\ \\ 
The work of Klaudia Bałazy was supported by the National Centre of Science (Poland) Grant No. 2020/39/D/ST6/ 01332. Klaudia Bałazy is affiliated with Doctoral School of Exact and Natural Sciences at the Jagiellonian University.
}

\bibliography{custom}


\appendix

\section{Related Work}

\paragraph{PEFT:} As large language models continue to grow, parameter-efficient fine-tuning (PEFT) has become a popular approach to reducing computational and storage costs. Among various methods~\cite{houlsby2019parameter,guo-etal-2021-parameter,li2021prefix,lester2021power}, LoRA~\cite{hu2021lora} has emerged as one of the most widely used. 
Building on its success, several approaches have been proposed to enhance different aspects of PEFT~\cite{kopiczko2023vera,adalora,dettmers2024qlora}. One such method, LoRA-XS~\cite{balazy2024lora}, further optimizes parameter efficiency by enabling flexible control over the number of trainable parameters per adaptation module. 
\our{} reuses the idea of SVD-based projections to reduce the parameter space dimensionality. 

\paragraph{Bayesian LoRAs:} Standard LoRA~\cite{hu2021lora} does not account for uncertainty, making fine-tuned models susceptible to miscalibration. Then, Bayesian LoRA approaches integrate Bayesian inference techniques into LoRA to improve uncertainty estimation and generalization.

Several Bayesian LoRA methods have been proposed, each employing different Bayesian techniques to address these challenges.  SWAG-LoRA \cite{onal2024gaussian} combines Stochastic Weight Averaging-Gaussian (SWAG) with LoRA to enable approximate Bayesian inference, significantly improving model calibration and reducing overconfidence. Laplace-LoRA \cite{robeyns2024laplaceLora} applies a Laplace approximation to the posterior over LoRA parameters. Bella \cite{doan2025bayesianlowranklearningbella} introduces an approach that reduces the cost of Bayesian deep ensembles by applying multiple low-rank perturbations to a pre-trained model.
BLoB (Bayesian Low-Rank Adaptation by Backpropagation) \cite{wang2024blob} jointly learns both the mean and covariance of model parameters throughout the fine-tuning process using Variational Inference. B-LoRA \cite{meo2024bayesianlora} introduces a Bayesian perspective to both quantization and rank selection by using a prior distribution over these hyperparameters, optimizing model efficiency and reducing bit operations.

 The key challenge lies in balancing uncertainty modeling with parameter efficiency, as Bayesian inference typically increases both the number of trainable parameters and computational cost.
Despite their advantages, Bayesian LoRA methods face challenges related to increased parameter count and computational cost. One major issue is the higher storage and memory requirements, as Bayesian methods often require additional parameters to model uncertainty, particularly those involving covariance estimation, such as SWAG-LoRA. 
Scalability remains a concern for methods that explicitly model uncertainty across a large number of parameters.

\section{Scientific Artifacts Licenses}

Listed below are the licenses for the scientific artifacts used in this research. For complete information, please use the links below and refer to the original sources.

Scientific Artifacts: RoBERTa-large (MIT), MRPC (\href{https://aclanthology.org/I05-5002/}{Unknown}), RTE (\href{https://tac.nist.gov/publications/2009/additional.papers/RTE5_overview.proceedings.pdf}{Unknown}), CoLA (\href{https://arxiv.org/pdf/1805.12471}{Unknown}), SST-2 (\href{https://aclanthology.org/D13-1170/}{Unknown}), HuggingFace Transformers Library (Apache-2.0), SWAG-LoRA repository\footnote{\url{https://github.com/fortuinlab/swag-lora}} (\href{https://github.com/fortuinlab/swag-lora}{MIT}), LoRA-XS   repository\footnote{\url{https://github.com/MohammadrezaBanaei/LoRA-XS}} (\href{https://github.com/MohammadrezaBanaei/LoRA-XS}{Unknown}).

\section{Model Size And Budget}

\begin{itemize}
    \item RoBERTa-large: 355M parameters
    \item GPUs: RTX4090 and V100-SXM2-32GB, each run was performed on a single GPU
    \item GPU total time: $\approx$ 63 days
\end{itemize}


\section{Statistics For Data}
We followed the original GLUE train-validation split
\begin{itemize}
    \item MRPC - train: 3'668, val: 408
    \item RTE - train: 2'490, val: 277
    \item CoLA - train: 8'551, val: 1043
    \item SST2 - train: 67'349, val: 872
\end{itemize}

\section{Experimental Setup Details}



The study was conducted on a subset of the GLUE benchmark~\cite{wang2019}, specifically on RTE, MRPC, CoLA, and SST-2 tasks (with the original train-validation split), using RoBERTa-large~\cite{liu2019} checkpoints from the HuggingFace Transformers library~\cite{wolf2020}. 
For the RTE and MRPC tasks, we followed LoRA-XS and initialized LoRA-XS modules with weights fine-tuned on the MNLI task.
We integrated B-LoRA-XS/LoRA-XS modules into the Query, Value, Attention Output, and Output Fully Connected weight matrices in all transformer layers~\cite{NIPS2017_3f5ee243}, whereas due to budget limits, standard LoRA and SWAG-LoRA modules were added only to the Query and Value matrices. Note this is sufficient for SWAG-LoRA to achieve its best performance.

For each dataset, for the burn-in stage of training, we adopted hyperparameters from the LoRA-XS paper. These include: learning rate, batch size, AdamW optimizer~\cite{loshchilov2019}, linear scheduler with warm-up, dropout, and the LoRA scaling factor $\alpha$. For standard LoRA we followed the same setup, except for the learning rate, which was determined through grid search. 
Similarly, the SWAG starting epoch (e.g. 10 or 25) was selected through grid search. 
Based on the findings from SWAG-LoRA, we used a constant learning rate scheduler (SWALR) with warm-up. The SWAG learning rate was set to the maximum learning rate from the first (burn-in) phase of training. Unless stated otherwise, we used a low-rank covariance matrix approximation with the rank $k=10$. In all our experiments, SWAG estimation was applied exclusively to the LoRA modules, and SWAG predictions were consistently obtained with $S=15$ model samples.

\begin{table*}[htp!]
\centering
\begin{tabular}{lllrrrrrr}
\toprule
 &  &  & \multicolumn{2}{r}{Accuracy} & \multicolumn{2}{r}{ECE} & \multicolumn{2}{r}{NLL} \\
 &  &  & median & s.d. & median & s.d. & median & s.d. \\
method & $r$ & $k$ &  &  &  &  &  &  \\
\midrule
\multirow[t]{2}{*}{LoRA} & 2 & 10 & 0.865 & 0.004 & 0.116 & 0.044 & 0.652 & 0.215 \\
\cline{2-9}
 & 8 & 10 & 0.870 & 0.004 & 0.125 & 0.008 & 0.933 & 0.154 \\
\cline{1-9} \cline{2-9}
\multirow[t]{4}{*}{LoRA-SWAG} & \multirow[t]{2}{*}{2} & 2 & 0.870 & 0.003 & 0.086 & 0.013 & 0.390 & 0.018 \\
 &  & 10 & 0.870 & 0.002 & 0.049 & 0.013 & 0.365 & 0.015 \\
\cline{2-9}
 & \multirow[t]{2}{*}{8} & 2 & 0.875 & 0.005 & 0.078 & 0.016 & 0.384 & 0.012 \\
 &  & 10 & 0.874 & 0.004 & 0.037 & 0.008 & 0.351 & 0.009 \\
\cline{1-9} \cline{2-9}
\multirow[t]{4}{*}{LoRA-XS} & 2 & 10 & 0.822 & 0.009 & 0.065 & 0.023 & 0.451 & 0.022 \\
\cline{2-9}
 & 8 & 10 & 0.853 & 0.002 & 0.059 & 0.021 & 0.440 & 0.052 \\
\cline{2-9}
 & 16 & 10 & 0.859 & 0.003 & 0.096 & 0.016 & 0.516 & 0.067 \\
\cline{2-9}
 & 25 & 10 & 0.869 & 0.002 & 0.099 & 0.021 & 0.465 & 0.102 \\
\cline{1-9} \cline{2-9}
\multirow[t]{7}{*}{B-LoRA-XS} & \multirow[t]{2}{*}{2} & 5 & 0.822 & 0.002 & 0.040 & 0.009 & 0.412 & 0.003 \\
 &  & 10 & 0.822 & 0.005 & 0.036 & 0.017 & 0.422 & 0.016 \\
\cline{2-9}
 & 8 & 10 & 0.855 & 0.004 & 0.044 & 0.005 & 0.372 & 0.018 \\
\cline{2-9}
 & \multirow[t]{2}{*}{16} & 5 & 0.863 & 0.003 & 0.038 & 0.020 & 0.354 & 0.037 \\
 &  & 10 & 0.863 & 0.001 & 0.046 & 0.007 & 0.367 & 0.006 \\
\cline{2-9}
 & \multirow[t]{2}{*}{25} & 5 & 0.870 & 0.003 & 0.041 & 0.006 & 0.360 & 0.021 \\
 &  & 10 & 0.869 & 0.002 & 0.049 & 0.013 & 0.378 & 0.016 \\
\cline{1-9} \cline{2-9}
\bottomrule
\end{tabular}
\caption{Numeric values for CoLA dataset.}
\label{tab:num_cola}
\end{table*}

\begin{table*}[htp!]
\centering
\begin{tabular}{lllrrrrrr}
\toprule
 &  &  & \multicolumn{2}{r}{Accuracy} & \multicolumn{2}{r}{ECE} & \multicolumn{2}{r}{NLL} \\
 &  &  & median & s.d. & median & s.d. & median & s.d. \\
method & $r$ & $k$ &  &  &  &  &  &  \\
\midrule
\multirow[t]{2}{*}{LoRA} & 2 & 10 & 0.912 & 0.003 & 0.069 & 0.010 & 0.406 & 0.230 \\
\cline{2-9}
 & 8 & 10 & 0.912 & 0.005 & 0.086 & 0.006 & 0.727 & 0.165 \\
\cline{1-9} \cline{2-9}
\multirow[t]{4}{*}{LoRA-SWAG} & \multirow[t]{2}{*}{2} & 2 & 0.917 & 0.004 & 0.112 & 0.035 & 0.332 & 0.034 \\
 &  & 10 & 0.917 & 0.005 & 0.052 & 0.016 & 0.306 & 0.031 \\
\cline{2-9}
 & \multirow[t]{2}{*}{8} & 2 & 0.912 & 0.003 & 0.056 & 0.035 & 0.331 & 0.031 \\
 &  & 10 & 0.912 & 0.004 & 0.048 & 0.018 & 0.321 & 0.127 \\
\cline{1-9} \cline{2-9}
\multirow[t]{4}{*}{LoRA-XS} & 2 & 10 & 0.861 & 0.017 & 0.048 & 0.010 & 0.338 & 0.022 \\
\cline{2-9}
 & 8 & 10 & 0.886 & 0.007 & 0.078 & 0.023 & 0.355 & 0.105 \\
\cline{2-9}
 & 16 & 10 & 0.904 & 0.006 & 0.079 & 0.015 & 0.450 & 0.135 \\
\cline{2-9}
 & 25 & 10 & 0.904 & 0.008 & 0.088 & 0.015 & 0.560 & 0.176 \\
\cline{1-9} \cline{2-9}
\multirow[t]{7}{*}{B-LoRA-XS} & \multirow[t]{2}{*}{2} & 2 & 0.860 & 0.012 & 0.080 & 0.029 & 0.386 & 0.027 \\
 &  & 10 & 0.858 & 0.012 & 0.046 & 0.011 & 0.336 & 0.025 \\
\cline{2-9}
 & 8 & 10 & 0.890 & 0.004 & 0.043 & 0.014 & 0.304 & 0.023 \\
\cline{2-9}
 & \multirow[t]{2}{*}{16} & 2 & 0.912 & 0.003 & 0.037 & 0.010 & 0.270 & 0.030 \\
 &  & 10 & 0.909 & 0.007 & 0.047 & 0.007 & 0.301 & 0.044 \\
\cline{2-9}
 & \multirow[t]{2}{*}{25} & 2 & 0.917 & 0.005 & 0.036 & 0.011 & 0.268 & 0.020 \\
 &  & 10 & 0.909 & 0.005 & 0.049 & 0.004 & 0.312 & 0.013 \\
\cline{1-9} \cline{2-9}
\bottomrule
\end{tabular}
\caption{Numeric values for MPRC dataset.}
\label{tab:num_mrpc}
\end{table*}

\begin{table*}[htp!]
\centering
\begin{tabular}{lllrrrrrr}
\toprule
 &  &  & \multicolumn{2}{r}{Accuracy} & \multicolumn{2}{r}{ECE} & \multicolumn{2}{r}{NLL} \\
 &  &  & median & s.d. & median & s.d. & median & s.d. \\
method & $r$ & $k$ &  &  &  &  &  &  \\
\midrule
\multirow[t]{2}{*}{LoRA} & 2 & 10 & 0.874 & 0.008 & 0.123 & 0.007 & 1.264 & 0.239 \\
\cline{2-9}
 & 8 & 10 & 0.874 & 0.010 & 0.125 & 0.010 & 1.072 & 0.123 \\
\cline{1-9} \cline{2-9}
\multirow[t]{2}{*}{LoRA-SWAG} & 2 & 10 & 0.870 & 0.007 & 0.091 & 0.009 & 0.518 & 0.046 \\
\cline{2-9}
 & 8 & 10 & 0.877 & 0.011 & 0.078 & 0.009 & 0.388 & 0.039 \\
\cline{1-9} \cline{2-9}
\multirow[t]{4}{*}{LoRA-XS} & 2 & 10 & 0.632 & 0.043 & 0.126 & 0.046 & 0.730 & 0.052 \\
\cline{2-9}
 & 8 & 10 & 0.870 & 0.005 & 0.116 & 0.017 & 0.698 & 0.188 \\
\cline{2-9}
 & 16 & 10 & 0.884 & 0.007 & 0.097 & 0.013 & 0.644 & 0.123 \\
\cline{2-9}
 & 25 & 10 & 0.902 & 0.005 & 0.099 & 0.006 & 0.957 & 0.045 \\
\cline{1-9} \cline{2-9}
\multirow[t]{4}{*}{B-LoRA-XS} & 2 & 10 & 0.650 & 0.062 & 0.079 & 0.025 & 0.652 & 0.024 \\
\cline{2-9}
 & 8 & 10 & 0.877 & 0.003 & 0.083 & 0.004 & 0.465 & 0.029 \\
\cline{2-9}
 & 16 & 10 & 0.888 & 0.007 & 0.073 & 0.011 & 0.446 & 0.030 \\
\cline{2-9}
 & 25 & 10 & 0.892 & 0.005 & 0.076 & 0.022 & 0.510 & 0.239 \\
\cline{1-9} \cline{2-9}
\bottomrule
\end{tabular}
\caption{Numeric values for RTE dataset.}
\label{tab:num_rte}
\end{table*}

\begin{table*}[htp!]
\centering
\begin{tabular}{lllrrrrrr}
\toprule
 &  &  & \multicolumn{2}{r}{Accuracy} & \multicolumn{2}{r}{ECE} & \multicolumn{2}{r}{NLL} \\
 &  &  & median & s.d. & median & s.d. & median & s.d. \\
method & $r$ & $k$ &  &  &  &  &  &  \\
\midrule
\multirow[t]{2}{*}{LoRA} & 2 & 10 & 0.961 & 0.003 & 0.030 & 0.006 & 0.162 & 0.027 \\
\cline{2-9}
 & 8 & 10 & 0.962 & 0.002 & 0.032 & 0.004 & 0.198 & 0.025 \\
\cline{1-9} \cline{2-9}
\multirow[t]{2}{*}{LoRA-SWAG} & 2 & 10 & 0.956 & 0.003 & 0.020 & 0.069 & 0.145 & 0.068 \\
\cline{2-9}
 & 8 & 10 & 0.966 & 0.004 & 0.030 & 0.033 & 0.141 & 0.031 \\
\cline{1-9} \cline{2-9}
\multirow[t]{4}{*}{LoRA-XS} & 2 & 10 & 0.944 & 0.001 & 0.026 & 0.002 & 0.168 & 0.003 \\
\cline{2-9}
 & 8 & 10 & 0.953 & 0.002 & 0.034 & 0.005 & 0.175 & 0.011 \\
\cline{2-9}
 & 16 & 10 & 0.959 & 0.002 & 0.032 & 0.003 & 0.161 & 0.012 \\
\cline{2-9}
 & 25 & 10 & 0.958 & 0.003 & 0.032 & 0.005 & 0.160 & 0.021 \\
\cline{1-9} \cline{2-9}
\multirow[t]{4}{*}{B-LoRA-XS} & 2 & 10 & 0.945 & 0.002 & 0.025 & 0.003 & 0.163 & 0.003 \\
\cline{2-9}
 & 8 & 10 & 0.952 & 0.001 & 0.019 & 0.006 & 0.152 & 0.008 \\
\cline{2-9}
 & 16 & 10 & 0.958 & 0.001 & 0.025 & 0.006 & 0.147 & 0.014 \\
\cline{2-9}
 & 25 & 10 & 0.961 & 0.000 & 0.027 & 0.005 & 0.137 & 0.007 \\
\cline{1-9} \cline{2-9}
\bottomrule
\end{tabular}
\caption{Numeric values for SST-2 dataset.}
\label{tab:num_sst2}
\end{table*}

\begin{table*}[htp!]
\centering 
\begin{tabular}{lllrrrrrr}
\toprule
 &  &  & \multicolumn{2}{r}{Accuracy} & \multicolumn{2}{r}{ECE} & \multicolumn{2}{r}{NLL} \\
 &  &  & median & s.d. & median & s.d. & median & s.d. \\
method & $r$ & $k$ &  &  &  &  &  &  \\
\midrule
LoRA & 2 & - & 0.865 & 0.004 & 0.116 & 0.047 & 0.652 & 0.228 \\
\cline{1-9} \cline{2-9}
\multirow[t]{10}{*}{LoRA-SWAG} & \multirow[t]{5}{*}{2} & 0 & 0.859 & 0.002 & 0.117 & 0.007 & 0.675 & 0.067 \\
 &  & 2 & 0.870 & 0.003 & 0.086 & 0.013 & 0.390 & 0.018 \\
 &  & 5 & 0.869 & 0.002 & 0.047 & 0.012 & 0.362 & 0.010 \\
 &  & 10 & 0.870 & 0.003 & 0.049 & 0.014 & 0.365 & 0.016 \\
 &  & 20 & 0.870 & 0.005 & 0.039 & 0.003 & 0.373 & 0.018 \\
\cline{2-9}
 & \multirow[t]{5}{*}{8} & 0 & 0.864 & 0.006 & 0.122 & 0.008 & 0.773 & 0.064 \\
 &  & 2 & 0.875 & 0.005 & 0.078 & 0.016 & 0.384 & 0.012 \\
 &  & 5 & 0.873 & 0.005 & 0.051 & 0.008 & 0.364 & 0.013 \\
 &  & 10 & 0.874 & 0.004 & 0.037 & 0.009 & 0.351 & 0.010 \\
 &  & 20 & 0.876 & 0.005 & 0.032 & 0.019 & 0.360 & 0.034 \\
\cline{1-9} \cline{2-9}
LoRA-XS & 16 & - & 0.859 & 0.004 & 0.096 & 0.017 & 0.516 & 0.071 \\
\cline{1-9} \cline{2-9}
\multirow[t]{5}{*}{B-LoRA-XS} & \multirow[t]{5}{*}{16} & 0 & 0.865 & 0.004 & 0.068 & 0.008 & 0.396 & 0.021 \\
 &  & 2 & 0.864 & 0.007 & 0.047 & 0.016 & 0.372 & 0.031 \\
 &  & 5 & 0.863 & 0.003 & 0.038 & 0.020 & 0.354 & 0.037 \\
 &  & 10 & 0.863 & 0.001 & 0.046 & 0.008 & 0.367 & 0.006 \\
 &  & 20 & 0.861 & 0.003 & 0.051 & 0.005 & 0.360 & 0.015 \\
\cline{1-9} \cline{2-9}
\bottomrule
\end{tabular}
\caption{Covariance matrix rank $k$ analysis for CoLA dataset.}
\label{tab:cov_cola}
\end{table*}

\begin{table*}[htp!]
\centering
\begin{tabular}{lllrrrrrr}
\toprule
 &  &  & \multicolumn{2}{r}{Accuracy} & \multicolumn{2}{r}{ECE} & \multicolumn{2}{r}{NLL} \\
 &  &  & median & s.d. & median & s.d. & median & s.d. \\
method & $r$ & $k$ &  &  &  &  &  &  \\
\midrule
LoRA & 2 & - & 0.912 & 0.003 & 0.069 & 0.011 & 0.406 & 0.244 \\
\cline{1-9} \cline{2-9}
\multirow[t]{10}{*}{LoRA-SWAG} & \multirow[t]{5}{*}{2} & 0 & 0.904 & 0.003 & 0.089 & 0.003 & 0.628 & 0.077 \\
 &  & 2 & 0.914 & 0.004 & 0.135 & 0.045 & 0.340 & 0.040 \\
 &  & 5 & 0.914 & 0.007 & 0.048 & 0.005 & 0.294 & 0.028 \\
 &  & 10 & 0.917 & 0.005 & 0.052 & 0.017 & 0.306 & 0.033 \\
 &  & 20 & 0.914 & 0.005 & 0.051 & 0.015 & 0.306 & 0.086 \\
\cline{2-9}
 & \multirow[t]{5}{*}{8} & 0 & 0.907 & 0.005 & 0.094 & 0.005 & 0.759 & 0.123 \\
 &  & 2 & 0.912 & 0.003 & 0.056 & 0.035 & 0.331 & 0.031 \\
 &  & 5 & 0.914 & 0.003 & 0.040 & 0.019 & 0.283 & 0.103 \\
 &  & 10 & 0.917 & 0.004 & 0.051 & 0.017 & 0.328 & 0.128 \\
 &  & 20 & 0.909 & 0.006 & 0.053 & 0.018 & 0.314 & 0.177 \\
\cline{1-9} \cline{2-9}
LoRA-XS & 16 & - & 0.909 & 0.007 & 0.079 & 0.011 & 0.388 & 0.095 \\
\cline{1-9} \cline{2-9}
\multirow[t]{5}{*}{B-LoRA-XS} & \multirow[t]{5}{*}{16} & 0 & 0.907 & 0.006 & 0.078 & 0.009 & 0.426 & 0.037 \\
 &  & 2 & 0.912 & 0.003 & 0.037 & 0.010 & 0.270 & 0.030 \\
 &  & 5 & 0.909 & 0.009 & 0.051 & 0.010 & 0.304 & 0.032 \\
 &  & 10 & 0.909 & 0.004 & 0.050 & 0.008 & 0.325 & 0.035 \\
 &  & 20 & 0.912 & 0.004 & 0.042 & 0.010 & 0.318 & 0.014 \\
\cline{1-9} \cline{2-9}
\bottomrule
\end{tabular}
\caption{Covariance matrix rank $k$ analysis for MRPC dataset.}
\label{tab:cov_mrpc}
\end{table*}

\section{Numeric Results}
Tables~\ref{tab:num_cola}-\ref{tab:cov_mrpc} present exact numeric values for the plots presented in Figures~\ref{fig:lorar}~and~\ref{fig:covrank}.

\section{Acknowledgments}
We acknowledge the use of ChatGPT for grammar checking and generation of the initial version of the plotting code.

\end{document}